\documentclass{article}
\usepackage{amsmath,graphicx,mlspconf}
\usepackage{xcolor}

\usepackage[numbers,sort&compress]{natbib} 

\usepackage{siunitx}
\usepackage{booktabs}
\usepackage[hidelinks]{hyperref}
\usepackage{tabularx}
\newcolumntype{Y}{>{\centering\arraybackslash}X}

\copyrightnotice{U.S.\ Government work not protected by U.S.\ copyright}

\copyrightnotice{979-8-3503-2411-2/25/\$31.00 {\copyright}2025 Crown}

\copyrightnotice{979-8-3503-2411-2/25/\$31.00 {\copyright}2025 European Union}

\copyrightnotice{979-8-3503-2411-2/25/\$31.00 {\copyright}2025 IEEE}

\toappear{2025 IEEE International Workshop on Machine Learning for Signal Processing, Aug.\ 31-- Sep.\ 3, 2025, Istanbul, Turkey}

\title{Few-Shot Radar Signal Recognition through Self-Supervised Learning and Radio Frequency Domain Adaptation}
\name{Zi Huang$^{1,2}$, Simon Denman$^{1}$, Akila Pemasiri$^{1}$, Clinton Fookes$^{1}$, Terrence Martin$^{2}$}
\address{
    $^{1}$Queensland University of Technology, Brisbane, Australia\\
    $^{2}$Revolution Aerospace, Brisbane, Australia
}

\begin{document}

\maketitle

\begin{abstract}
Radar signal recognition (RSR) plays a pivotal role in electronic warfare (EW), as accurately classifying radar signals is critical for informing decision-making. Recent advances in deep learning have shown significant potential in improving RSR in domains with ample annotated data. However, these methods fall short in EW scenarios where annotated radio frequency (RF) data are scarce or impractical to obtain. To address these challenges, we introduce a self-supervised learning (SSL) method which utilises masked signal modelling and RF domain adaption to perform few-shot RSR and enhance performance in environments with limited RF samples and annotations. We propose a two-step approach, first pre-training masked autoencoders (MAE) on baseband in-phase and quadrature (I/Q) signals from diverse RF domains, and then transferring the learned representations to the radar domain, where annotated data are scarce. Empirical results show that our lightweight self-supervised ResNet1D model with domain adaptation achieves up to a 17.5\% improvement in $1$-shot classification accuracy when pre-trained on in-domain signals (i.e., radar signals) and up to a 16.31\% improvement when pre-trained on out-of-domain signals (i.e., comm signals), compared to its baseline without using SSL. We also present reference results for several MAE designs and pre-training strategies, establishing a new benchmark for few-shot radar signal classification.
\end{abstract}

\section{Introduction}
\label{sec:intro}
\vspace{-5pt}

Radar signal recognition (RSR) is a crucial capability in cognitive electronic warfare (EW) \cite{haigh_cognitive_2021-1}, where accurate radar signal classification is essential for informed decision-making in the battlefield. Recent progress in deep learning has demonstrated significant potential \cite{geng2021deep} in addressing RSR sub-tasks, such as automatic modulation classification (AMC) \cite{oshea_over--air_2018} and radar activity segmentation \cite{huang2024multi}, when abundant radio frequency (RF) data are available for model development. However, most existing RSR methods rely heavily on annotated RF data, which presents a challenge in EW scenarios where mission-specific data are often scarce or difficult to acquire. Moreover, practical EW operations often demand rapid in-mission re-training of models on limited data to respond to the rapidly changing threat landscape \cite{haigh_cognitive_2021-1}. This makes few-shot learning a critical yet underexplored area in RSR.


Supervised RF signal recognition has been an active research area over the past decade \cite{oshea_over--air_2018, vila2019deep, huynh-the_automatic_2021, caromi2021deep, jagannath_multi-task_2021-1, clerico2023lstm, huang2023multi, pemasiri2024automatic}. Previous methods have relied on 2D feature transformations \cite{o2017learning, huynh2021accurate, logue2019expert} to enhance classification performance in supervised settings. More recently, 1D approaches \cite{clerico2023lstm, huang2023multi, tian2024novel} have shown improved performance by capturing fine-grained temporal relationships in I/Q signals, as demonstrated in various signal recognition tasks within the RF domain, such as modulation classification \cite{shi2019deep, jagannath_multi-task_2021-1}, radar signal characterisation \cite{huang2023multi}, RF fingerprinting \cite{zhang2024few}, and radar activity segmentation \cite{huang2024multi}. While unsupervised methods \cite{oshea_unsupervised_2016-1, ali2017unsupervised} have been applied to AMC, self-supervised approaches have been more extensively studied in related signal processing domains, such as visual recognition \cite{bao2021beit, he2022masked}, audio recognition \cite{huang2022masked, gong2022ssast}, bio-signal classification \cite{chien2022maeeg, liu2023frequency}, and natural language processing \cite{kenton2019bert}. Notably, masked autoencoders (MAE) have proven to be effective self-supervised learning (SSL) architectures for time series classification \cite{zerveas2021transformer} and representation learning in related tasks \cite{xie2022simmim, chien2022maeeg, gong2022ssast, bai2023masked}. Due to their success, MAE-based methods have been adapted in recent work \cite{huang2022deep, yao2023few} for RF emitter identification, particularly when training samples are limited. However, self-supervised approaches for RSR remain largely unexplored, likely due to the scarcity of publicly available radar datasets \cite{geng2021deep, huang2023multi}. Moreover, existing work in RSR has yet to explore SSL with RF domain adaptation.


In this work, we introduce a flexible two-step SSL approach that leverages masked signal modelling (MSM) and domain adaptation for few-shot RSR. Our proposed method pre-trains models on baseband I/Q signals with diverse characteristics from various RF domains. We then apply few-shot transfer learning to adapt pre-trained models to the radar domain using limited training samples. Our main contributions are as follows: (i) we propose a modular and effective SSL architecture using different MSM strategies to enhance radar signal classification performance in few-shot settings; (ii) we introduce RF domain adaptation for self-supervised pre-training, enabling effective adaptation to the radar domain with limited data; and (iii) we release our SSL datasets\footnote{The datasets used for pre-training, fine-tuning, and evaluation are available at: \url{https://github.com/abcxyzi/RadCharSSL}} accompanied by a new benchmark for few-shot radar signal classification. To the best of our knowledge, the integration of domain adaptation of RF signals with varying sequence lengths and sampling frequencies from diverse RF domains into the radar domain has not been explored previously.

\vspace{-5pt}
\section{Proposed Method}
\label{sec:method}
\vspace{-5pt}

\subsection{Two-Step Few-Shot Learning}
\begin{figure}[tb]
    \centering
    \includegraphics[width=1\linewidth]{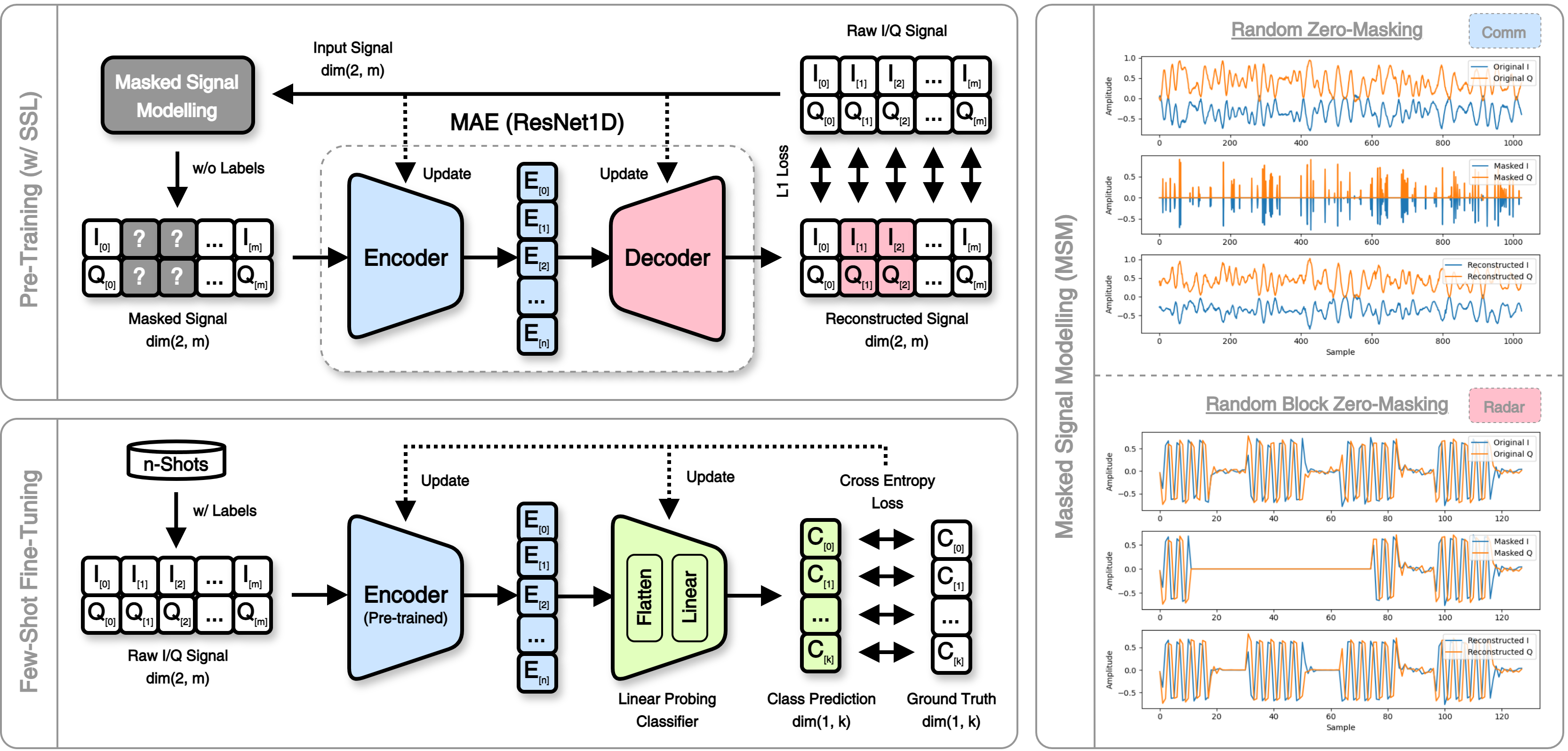}
    \caption{Proposed two-step SSL architecture. The encoder is pre-trained on various RF domains using the most optimal masking strategy, and fine-tuning is conducted on in-domain radar data using a linear probing classifier.}
\label{fig:model_arch}
\vspace{-5pt}
\end{figure}

\vspace{-5pt}
Our proposed SSL approach comprises two sequential steps. First, annotation-free pre-training of a masked autoencoder is conducted on a source RF domain (i.e., radar, comm, or a mixture of both). Then, the pre-trained encoder is fine-tuned on the target radar domain using a limited amount of annotated data. We follow the modular encoder-decoder paradigm \cite{hinton2006reducing} to construct our autoencoder as shown in Fig. \ref{fig:model_arch}. In the pre-training step, we utilise asymmetric masked autoencoding \cite{he2022masked} whereby the model operates on a partially observed I/Q signal in the presence of additive white Gaussian noise (AWGN). We pre-train the model to reconstruct the original signal using a sample-wise similarity loss function \cite{huang2024multi}. This process is considered self-supervised as no annotations are required for the reconstruction task. After pre-training, we replace the decoder with a linear probing classifier consisting of a flatten operation followed by a simple fully connected layer. We fine-tune the classifier on the target domain with a small amount (i.e., a few shots) of annotated RF data before evaluating it on a large test set. 

\vspace{-5pt}
\subsection{Lightweight Autoencoders for I/Q Signals}
\vspace{-5pt}
We implement several autoencoders to examine their effectiveness in self-supervised RSR. Specifically, we focus our experiments on lightweight models, prioritising rapid re-training and computational efficiency as they are critical considerations for adaptability in EW \cite{haigh_cognitive_2021-1}. We implement a ResNet1D \cite{huang2022deep}, a two-stage MS-TCN \cite{farha2019ms, huang2024multi}, and WaveNet \cite{van2016wavenet, jayashankar2024data} as our autoencoders. Notably, being a well established baseline, WaveNet \cite{jayashankar2024data} demonstrates exceptional performance in RF signal reconstruction, surpassing more complex and recent transformer-based models \cite{jayashankar2024data}. In our experiments, pre-training of MS-TCN and WaveNet are conducted in an end-to-end manner as they do not have independent encoders and decoders. As such, the linear probing classifier is appended to these models during fine-tuning. For our SSL benchmark, we use the $\ell_1$ regression loss to compute sample-wise similarity during pre-training, and the categorical cross-entropy loss for fine-tuning the classifier.

\subsection{Masked Signal Modelling}
\vspace{-5pt}
Masked signal modelling (MSM) is conceptually similar to masked image modelling (MIM) \cite{chen2020generative, xie2022simmim}, which has gained popularity as a simple and effective SSL approach in the computer vision domain. Our approach utilises the proxy task of I/Q signal reconstruction, training a model to reconstruct intentionally corrupted signals. By applying this process across a large corpus of examples, the model learns the salient representations necessary for accurate signal reconstruction. Similar to MIM, MSM involves several key design considerations, including the masking strategy $S_{\text{m}}$, masking ratio $R_{\text{m}}$, and model design. While the choice of model design is often influenced by external factors such as computational and memory requirements, $S_{\text{m}}$ and $R_{\text{m}}$ have a more explicit impact on the quality of pre-training \cite{xie2022simmim}.

We introduce several masking strategies for MSM, including random zero-masking (strat. A), random block zero-masking (strat. B), random noise-masking (strat. C), and block noise-masking (strat. D). Random masking involves modifying I/Q samples with mask values (i.e., either zero or noise), while block masking randomly obscures a continuous segment of the I/Q sequence (analogous to intentional interference in an EW setting). The extent of masking is governed by $R_{\text{m}}$, with higher values masking a larger percentage of the signal and thus making the reconstruction task more challenging. As shown by MIM \cite{xie2022simmim}, random masking promotes the learning of local relationships between samples, while block masking promotes the learning of broader, global patterns from the data. Our signal masking process is formalised as
\begin{equation}
s_{\text{m}} =
    \begin{cases}
    s \otimes 0_{\{X < R_{\text{m}}\}} & \text{if zero-masking}, \\
    s \oplus \hat{n}_{\{X < R_{\text{m}}\}} & \text{if noise-masking}, \\
    s & \text{if no masking},
    \end{cases}
\label{eq:masking}
\end{equation}
\begin{equation}
\hat{n} \sim \mathcal{N}(\mu_{\text{train}}, \sigma_{\text{train}}^2), \quad X \sim \mathcal{U}(0, 1), \quad R_{\text{m}} \in [0, 1],
\label{eq:sampling}
\end{equation}
\noindent where $s$ and $s_{\text{m}}$ denote the original and masked I/Q sequences respectively, $\otimes$ and $\oplus$ denote sample-wise multiplication and addition respectively, and $\hat{n}$ denotes random noise sampled using statistics computed from the training data distribution of the source domain.

\vspace{-5pt}
\subsection{RF Datasets for Domain Adaption}
\vspace{-5pt}
We explore domain adaptation by applying MSM to I/Q signals from diverse RF domains, including telecommunications and radar signals. Our approach involves pre-training, without annotations, MAEs on four diverse RF datasets: RadioML \cite{oshea_over--air_2018}, DeepRadar \cite{clerico2023lstm}, RadarComm \cite{jagannath_multi-task_2021-1}, and RadChar-SSL \cite{huang2023multi}. We then perform few-shot transfer learning to fine-tune each pre-trained model on RadChar-$n$Shot and evaluate performance on a separate test set (RadChar-Eval) derived from RadChar \cite{huang2023multi}. Here, $n$ represents the number of shots where $1$-shot corresponds to precisely $205$ frames (i.e., representing $1$ unique training example per signal class and SNR level), while $10$-shot consists of $2,050$ frames. We use only $10\%$ of RadioML for pre-training to ensure comparable dataset sizes across all pre-training datasets. Table \ref{table:datasets} provides a summary of the key characteristics of each dataset.

\begin{table}[t]
\vspace{-8pt}
\caption{Summary of RF signal classification datasets.}
\label{table:datasets}
\small 
\def\arraystretch{0.75}
\noindent\begin{tabularx}{\columnwidth}{>{\raggedright\arraybackslash}l|l|c|c|c|c}
    \toprule
    \textbf{Dataset}
        & \textbf{Type}
        & \textbf{Frames} 
        & \textbf{Size}
        & \textbf{$N_\text{cls}$}
        & \textbf{$t_\text{res}$ ($\si{\micro\second}$)} \\
    \midrule
    RadioML \cite{oshea_over--air_2018}
        & Comm
        & $255,590$
        & $1024$
        & $24$ 
        & $1$ \\
    DeepRadar \cite{clerico2023lstm}
        & Radar
        & $782,000$
        & $1024$
        & $23$
        & $0.01$ \\
    RadarComm \cite{jagannath_multi-task_2021-1}
        & Mixed
        & $125,361$
        & $128$
        & $6$ 
        & $0.1$ \\
    RadChar-SSL
        & Radar
        & $500,000$
        & $512$
        & $5$
        & $0.3$ \\   
    RadChar-$n$Shot
        & Radar
        & $205n$
        & $512$
        & $5$
        & $0.3$ \\
    RadChar-Eval
        & Radar
        & $100,000$
        & $512$
        & $5$
        & $0.3$ \\    
    \bottomrule
\end{tabularx}
\end{table}

\vspace{-5pt}
\section{Experiments}
\label{sec:experiments}
\vspace{-5pt}

\subsection{Training Details}
\vspace{-5pt}
We perform pre-training, fine-tuning, and model evaluation on a single Nvidia Tesla A100 GPU. All models are trained with the Adam optimiser, where constant learning rates of $0.001$ and $0.0001$ are used for self-supervised pre-training and fine-tuning, respectively. For pre-training, we train each model for $100$ epochs with early stopping based on validation loss with a $3$-step patience, using a fixed $70$-$20$-$10$ train-validation-test split for each dataset. For fine-tuning on limited frames, we train each model for the full $100$ epochs with the pre-trained encoder weights frozen for the first $10$ epochs. To maintain an even class distribution, no validation split is considered during fine-tuning. We establish baseline few-shot performance for each model using the same model configuration as that used in fine-tuning, but without the pre-trained weight initialisation. A fixed batch size of $128$ is used for pre-training, while a batch size of $8$ is used for both fine-tuning and baseline training. To improve generalisation and training stability, we standardise input signals with the mean and variance sampled from the training set of the respective RF domain. All models are evaluated on the RadChar-Eval dataset.

\subsection{Few-Shot Radar Signal Recognition}
\label{ssec:discussion}
\vspace{-5pt}
We establish a novel benchmark for few-shot RSR through the following experiments: (i) evaluating different MAEs; (ii) evaluating the impact of SSL on test performance across different SNR settings; (iii) evaluating different masking strategies; and (iv) evaluating the effectiveness of self-supervised pre-training on signals from different RF domains for few-shot signal classification. We evaluate the test performance of each linear probing classifier using classification accuracy and the macro F1 score. The test performance of each classifier is compared to its baseline (w/o SSL), where the model is trained from scratch without self-supervised pre-training.

\begin{table*}[tb]
\caption{Comparison of optimal few-shot radar classification performance results on the RadChar-Eval dataset.}
\label{table:massive_table}
\small 
\def\arraystretch{0.75} 
\begin{tabularx}{\textwidth}{>{\raggedright\arraybackslash}p{1.2cm}@{\hspace{1em}}|l|Y|Y|c|c|Y|Y|c|c|Y|Y|c|c}
    \toprule
    \textbf{Model} & \textbf{Pre-Training} & \multicolumn{4}{c|}{\textbf{1-Shot Eval. (Best)}} & \multicolumn{4}{c|}{\textbf{5-Shot Eval. (Best)}} & \multicolumn{4}{c}{\textbf{10-Shot Eval. (Best)}} \\
    & \textbf{w/ SSL} & 
    \textbf{$S_{\text{m}}$} & \textbf{$R_{\text{m}}$} & \textbf{Acc. ($\%$)} & \textbf{F1 ($\%$)} 
    & \textbf{$S_{\text{m}}$} & \textbf{$R_{\text{m}}$} & \textbf{Acc. ($\%$)} & \textbf{F1 ($\%$)} 
    & \textbf{$S_{\text{m}}$} & \textbf{$R_{\text{m}}$} & \textbf{Acc. ($\%$)} & \textbf{F1 ($\%$)} \\
    \midrule

    ResNet1D & RadChar-SSL & A & 0.7 & $\mathbf{75.06}$ & $\mathbf{72.32}$ & A & 0.7 & $\mathbf{79.76}$ & $\mathbf{77.76}$ & A & 0.8 & $\mathbf{79.86}$ & $\mathbf{77.73}$ \\
    & DeepRadar & A & 0.4 & $68.32$ & $65.35$ & B & 0.2 & $76.15$ & $73.91$ & B & 0.2 & $76.87$ & $74.59$ \\
    & RadarComm & D & 0.7 & $70.03$ & $66.98$ & C & 0.8 & $77.21$ & $75.00$ & B & 0.7 & $78.05$ & $75.82$ \\
    & RadioML & A & 0.1 & $74.30$ & $71.81$ & A & 0.1 & $77.35$ & $75.16$ & B & 0.7 & $78.08$ & $75.96$ \\   
    & None & - & - & $63.88$ & $60.50$ & - & - & $75.02$ & $72.66$ & - & - & $76.57$ & $74.20$ \\
    \midrule
    
    MS-TCN & RadChar-SSL & A & 0.7 & $\mathbf{72.62}$ & $\mathbf{70.13}$ & A & 0.9 & $\mathbf{81.12}$ & $\mathbf{79.27}$ & A & 0.9 & $\mathbf{81.27}$ & $\mathbf{79.40}$ \\
    & DeepRadar & C & 0.6 & $52.73$ & $50.14$ & C & 0.6 & $70.29$ & $67.66$ & C & 0.7 & $74.86$ & $72.53$ \\
    & RadarComm & A & 0.9 & $68.40$ & $65.72$ & B & 0.7 & $76.13$ & $73.89$ & B & 0.9 & $78.25$ & $76.14$ \\
    & RadioML & A & 0.8 & $64.76$ & $61.95$ & A & 0.7 & $75.36$ & $72.96$ & A & 0.8 & $77.89$ & $75.75$ \\
    & None & - & - & $44.00$ & $40.81$ & - & - & $66.49$ & $63.69$ & - & - & $76.47$ & $74.17$ \\
    \midrule
    
    WaveNet & RadChar-SSL & A & 0.6 & $\mathbf{73.07}$ & $\mathbf{70.63}$ & A & 0.8 & $\mathbf{82.00}$ & $\mathbf{80.26}$ & 
    A & 0.8 & $\mathbf{83.37}$ & $\mathbf{81.64}$ \\
    & DeepRadar & D & 0.2 & $62.64$ & $59.86$ & A & 0.8 & $73.80$ & $71.30$ & A & 0.8 & $76.46$ & $74.25$ \\
    & RadarComm & D & 0.8 & $67.09$ & $64.43$ & B & 0.9 & $75.79$ & $73.53$ & B & 0.9 & $77.83$ & $75.70$ \\
    & RadioML & B & 0.9 & $64.84$ & $61.62$ & B & 0.9 & $77.40$ & $75.24$ & A & 0.9 & $78.66$ & $76.57$ \\        
    & None & - & - & $64.04$ & $60.93$ & - & - & $75.89$ & $73.55$ & - & - & $77.21$ & $75.10$ \\

    \bottomrule
\end{tabularx}
\end{table*}

Table \ref{table:massive_table} presents a summary of our results. The classification performance corresponding to the optimal selection of $S_{\text{m}}$ and $R_{\text{m}}$ used during pre-training is shown for each model. Here, we observe that fine-tuned models that are pre-trained on data from the same RF domain (i.e., RadChar) yield the greatest improvement in test performance when compared to their respective baselines. This performance gain is most substantial in the $1$-shot setting, which is the most challenging with the fewest labels available. The improvement diminishes as the number of annotated frames used in fine-tuning increases, as reflected by the $17.5\%$, $6.3\%$, and $4.3\%$ increases in $1$-shot, $5$-shot, and $10$-shot performance for ResNet1D, respectively. Separately, we observe that MS-TCN's baseline performance is relatively low in the $1$-shot setting, likely due to its larger model size ($33.6$ million parameters). This model requires more fine-tuning samples (Table \ref{table:massive_table}) to achieve a similar performance compared to ResNet1D ($2.7$ million parameters) and WaveNet ($1$ million parameters).


\begin{figure}[tb]
\begin{minipage}[b]{0.495\linewidth}
  \centering
  \centerline{\includegraphics[width=1\linewidth]{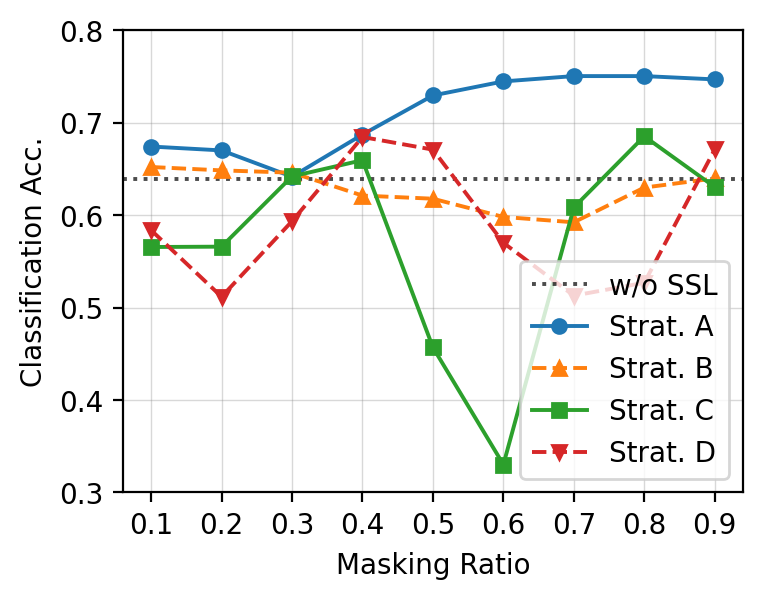}}
  \centerline{(a) RadChar-SSL (Radar)}\medskip
\end{minipage}
\hfill
\begin{minipage}[b]{0.495\linewidth}
  \centering
  \centerline{\includegraphics[width=1\linewidth]{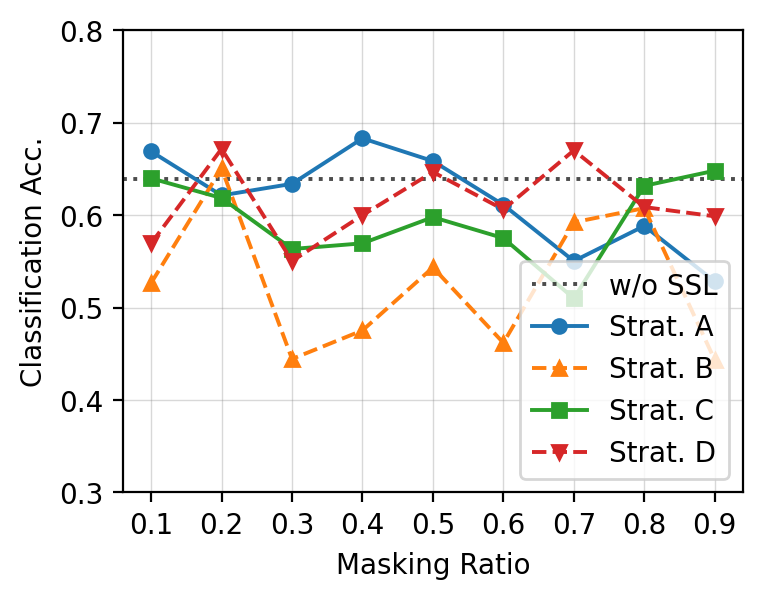}}
  \centerline{(b) DeepRadar (Radar)}\medskip
\end{minipage}
\hfill
\begin{minipage}[b]{0.495\linewidth}
  \centering
  \centerline{\includegraphics[width=1\linewidth]{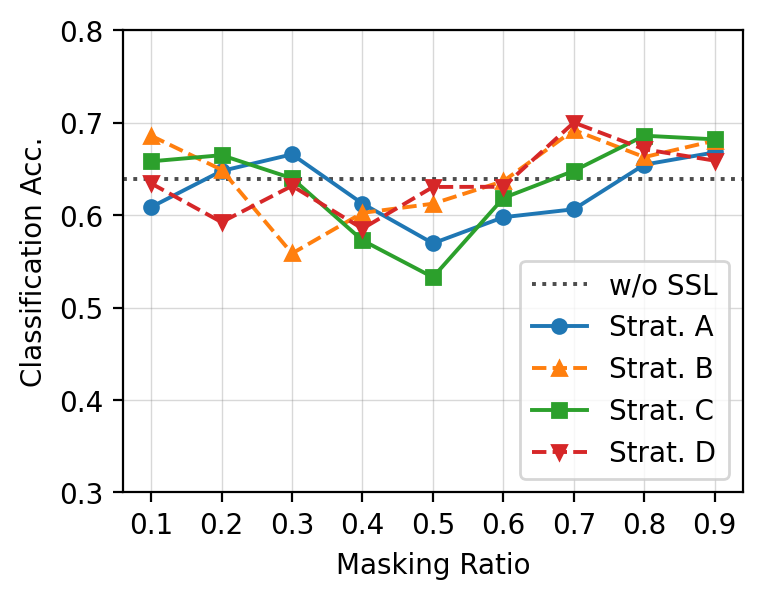}}
  \centerline{(c) RadarComm (Mixed)}\medskip
\end{minipage}
\hfill
\begin{minipage}[b]{0.495\linewidth}
  \centering
  \centerline{\includegraphics[width=1\linewidth]{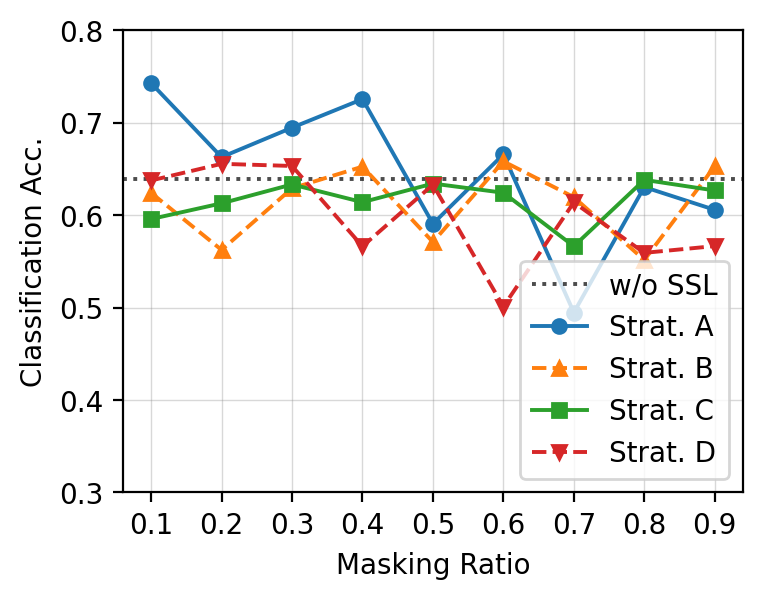}}
  \centerline{(d) RadioML (Comm)}\medskip
\end{minipage}
\hfill
\vspace{-22pt}
\caption{Impact of masking strategies on the $1$-shot classification performance of ResNet1D, pre-trained across different RF domains (a)–(d), and evaluated on RadChar-Eval.}
\label{fig:quad_mrs}
\vspace{-5pt}
\end{figure}

\begin{figure}[ht]
\begin{minipage}[b]{0.495\linewidth}
  \centering
  \centerline{\includegraphics[width=1\linewidth]{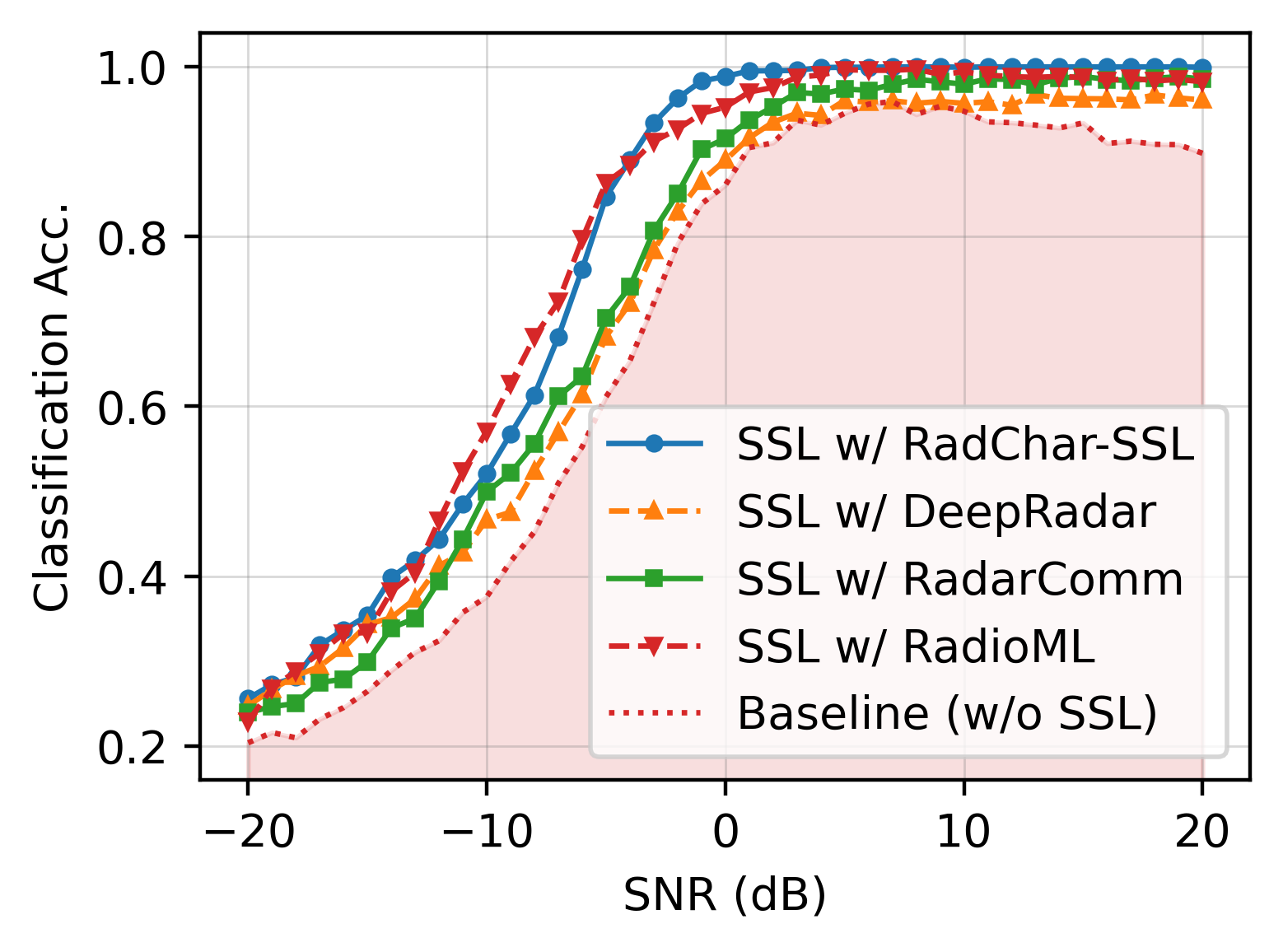}}
  \centerline{(a) ResNet ($n=1$)}\medskip
\end{minipage}
\hfill
\begin{minipage}[b]{0.495\linewidth}
  \centering
  \centerline{\includegraphics[width=1\linewidth]{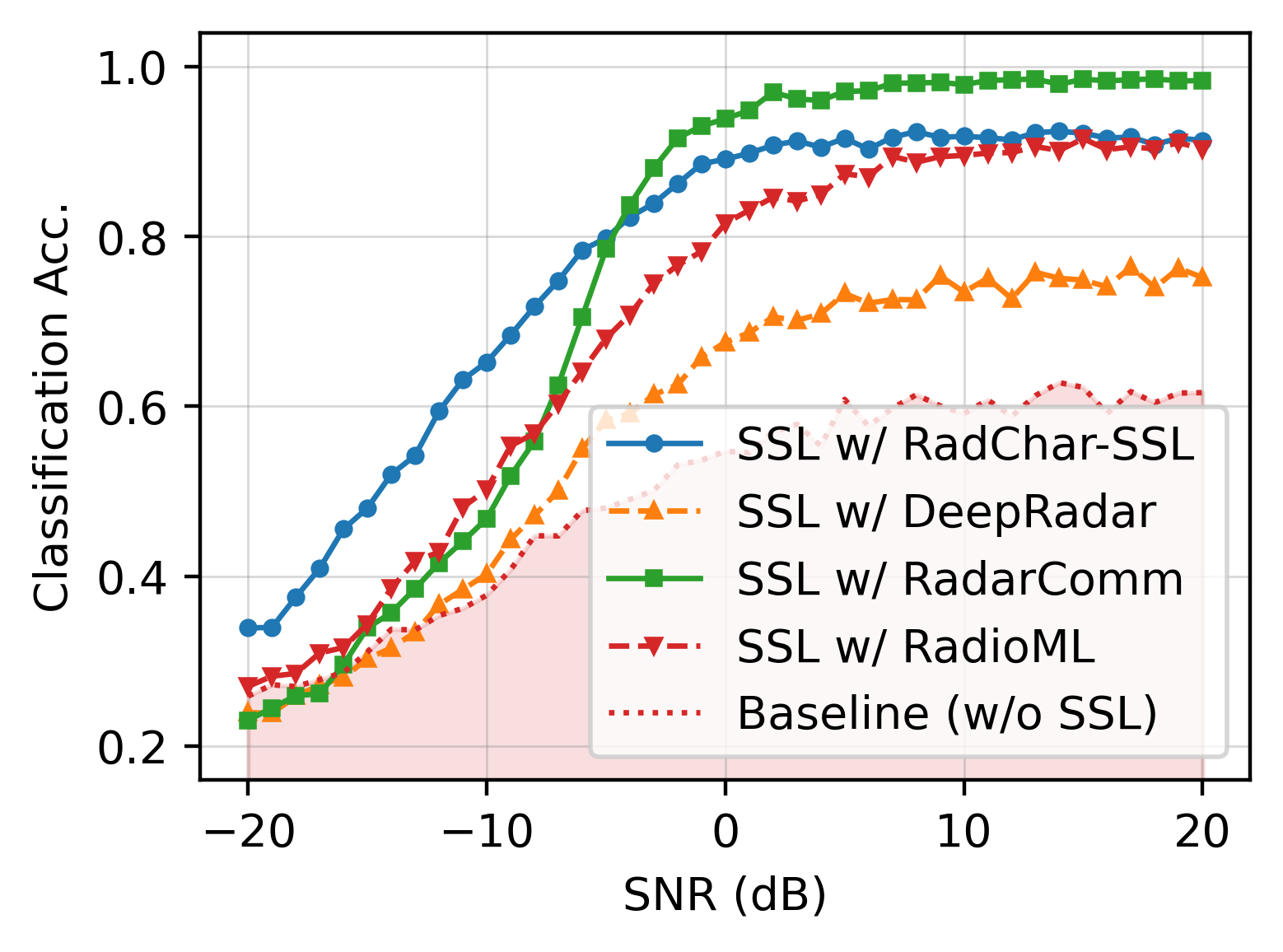}}
  \centerline{(b) MS-TCN ($n=1$)}\medskip
\end{minipage}
\hfill
\begin{minipage}[b]{0.495\linewidth}
  \centering
  \centerline{\includegraphics[width=1\linewidth]{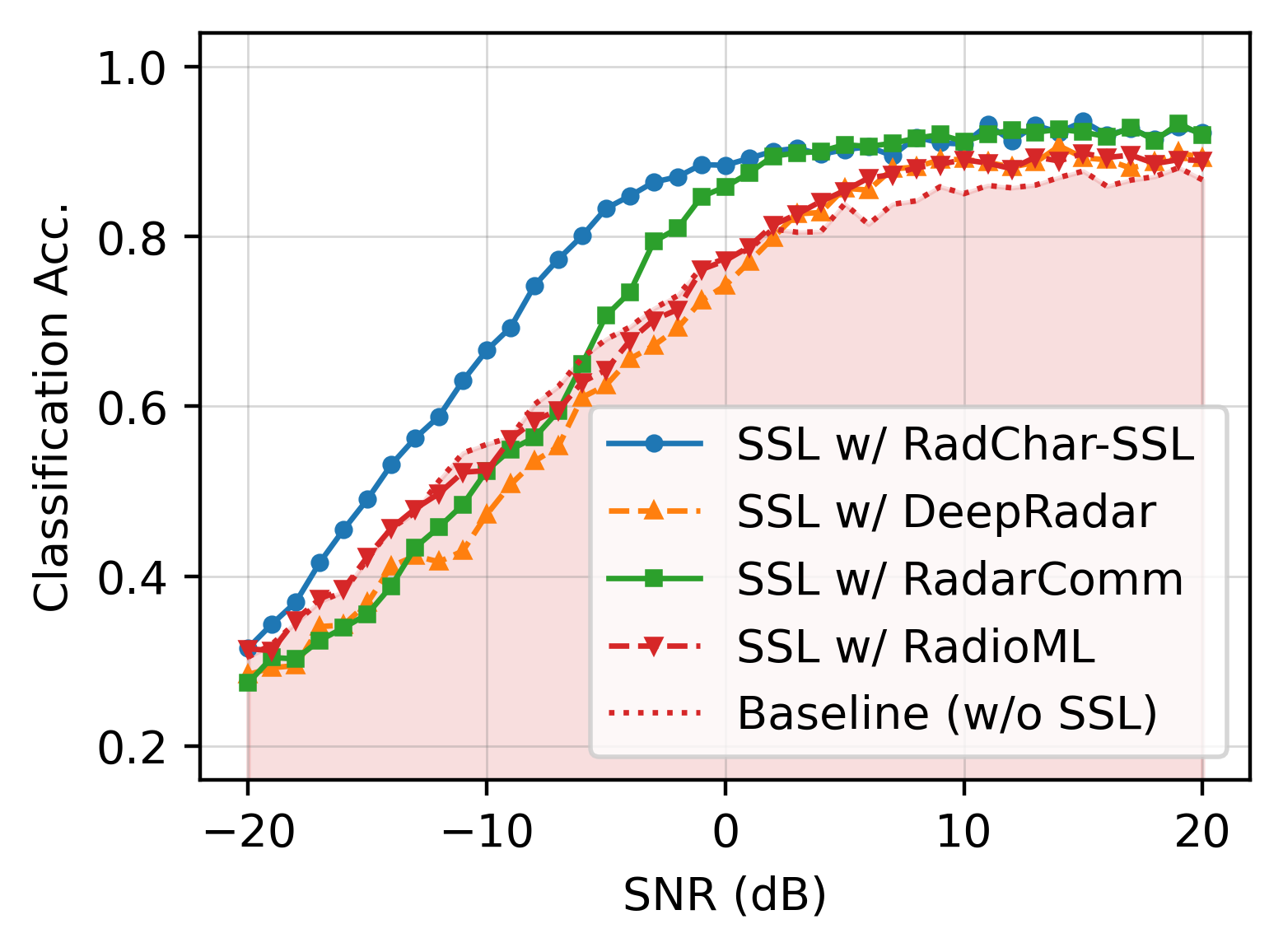}}
  \centerline{(c) WaveNet ($n=1$)}\medskip
\end{minipage}
\hfill
\begin{minipage}[b]{0.495\linewidth}
  \centering
  \centerline{\includegraphics[width=1\linewidth]{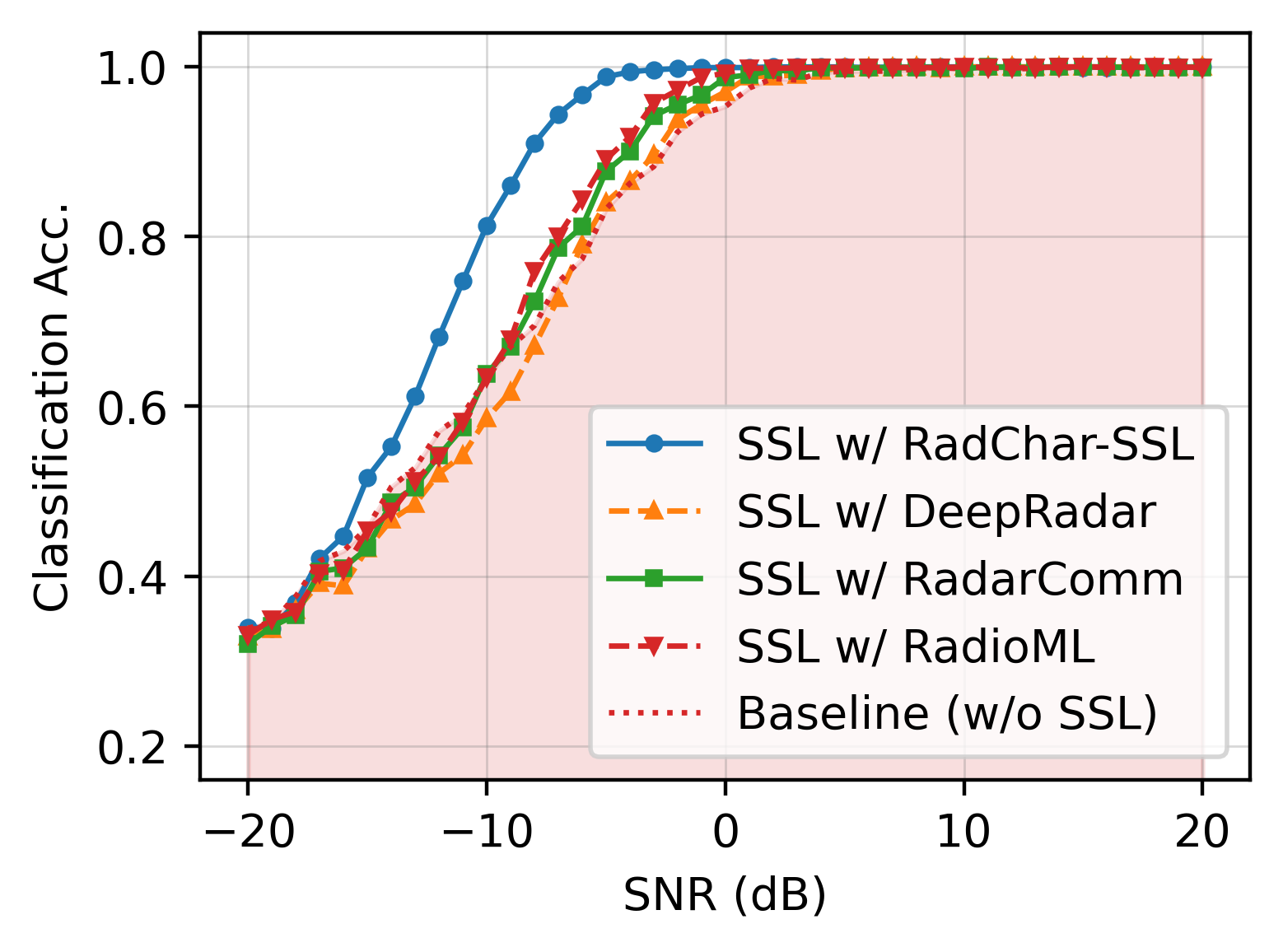}}
  \centerline{(d) WaveNet ($n=10$)}\medskip
\end{minipage}
\hfill
\vspace{-22pt}
\caption{$n$-shot classification performance of different models on RadChar-Eval. The optimal masking technique ($S_{\text{m}}$, $R_{\text{m}}$) for each pre-training domain is given in Table~\ref{table:massive_table}.}
\label{fig:quad_snrs}
\vspace{-12pt}
\end{figure}



The benefit of SSL on model performance is highly dependent on SNR. Notably, models tend to benefit the most from pre-training when evaluating in moderate to high SNR levels (Fig. \ref{fig:quad_snrs}). We hypothesise that signal features (e.g., number of pulses, pulse width) become more distinguishable at higher SNRs, enabling models to learn salient information through reconstruction. In contrast, at lower SNRs, random noise dominates, making reconstruction more challenging. In the most demanding fine-tuning configuration (i.e., $1$-shot), ResNet1D shows a performance boost ranging from $11.3\%$ to $25.7\%$ as SNR increases from $-20$ to $20$ \si{\deci\bel} when pre-trained on RadChar-SSL, whereas WaveNet's improvement is more modest ($3.5\%$ to $6.4\%$) when compared to their respective baselines without SSL. WaveNet employs dilated convolutions, which provide a larger receptive field compared to ResNet1D. Consequently, WaveNet may benefit less from MSM, as the masking process may not present enough of a challenge for reconstruction. This is reflected in WaveNet's preference for higher masking ratios (Table \ref{table:massive_table}). Our results (Table \ref{table:massive_table}) also suggest that classification performance saturates at around $10$ shots for both in-domain and out-of-domain fine-tuning. For instance, the classification accuracy of the in-domain pre-trained WaveNet improves from $73.07\%$ ($1$-shot) to $82\%$ ($5$-shot). However, performance improves only marginally from $82\%$ to $83.37\%$ between the $5$-shot and $10$-shot setting, respectively. These findings are consistent with a related study on SSL for RF fingerprinting of five unique emitters \cite{huang2022deep}, which reported test accuracy saturation at approximately $2,500$ frames on a private RF dataset.


The optimal masking strategy for classification performance depends on the model and the pre-training domain. For ResNet1D (Fig. \ref{fig:quad_mrs}), random zero-masking (strat. A) with lower masking ratios (below $0.4$) performs best when pre-training on RadioML and DeepRadar, while the same strategy with a higher masking ratio (above $0.7$) is more beneficial when pre-training on RadChar-SSL. Random block noise-masking (strat. D) with a higher masking ratio (above $0.7$) performs best when pre-training on the mixed RF domain RadarComm. While self-supervised pre-training on out-of-domain RF data using the optimal masking strategy improves performance across various few-shot settings (Table \ref{table:massive_table}), the effectiveness of RF domain adaptation depends on several key factors, particularly the extent of the domain shift between datasets. We hypothesise that differences in sampling rates (i.e., temporal resolution) between datasets affect how models learn the temporal relationships in the I/Q sequence. Pre-training on signals (Table \ref{table:datasets}) with higher temporal resolution ($t_\text{res}$) may lead to the model learning features that are less relevant when fine-tuned on signals with lower sampling rates, and vice versa. While resampling the source domain to match the target domain is possible, it reduces signal information and can introduce aliasing, especially when the original signal has limited I/Q samples, such as in RadarComm (Table \ref{table:datasets}). Conversely, upsampling requires interpolation, which may not accurately capture the characteristics of the original signal, potentially leading to bias and overfitting. 

Differences in domain-specific characteristics and the number of signal classes ($N_\text{cls}$) in the source and target domains (Table \ref{table:datasets}) can have a significant impact on fine-tuning performance. This is highlighted by comparing in-domain results (RadChar-SSL) with out-of-domain results in Fig. \ref{fig:quad_snrs}. Performance improvements from SSL are most notable for RadChar-SSL and RadarComm, where the number of signal classes ($6$ and $5$) and temporal resolution ($0.1$ and $0.3$) are most similar (Table \ref{table:datasets}). In contrast, DeepRadar and RadioML contain substantially more RF classes than RadChar. Furthermore, the signal features from DeepRadar and RadioML deviate significantly from the target domain. For example, the temporally smooth telecommunications signals from RadioML and the challenging low probability of intercept (LPI) radar classes (e.g., Px codes) in DeepRadar do not appear in the target domain, which may lead to over-regularisation and the reduced effectiveness of fine-tuning. 

\begin{figure}[ht]
\begin{minipage}[b]{0.495\linewidth}
  \centering
  \centerline{\includegraphics[width=1\linewidth]{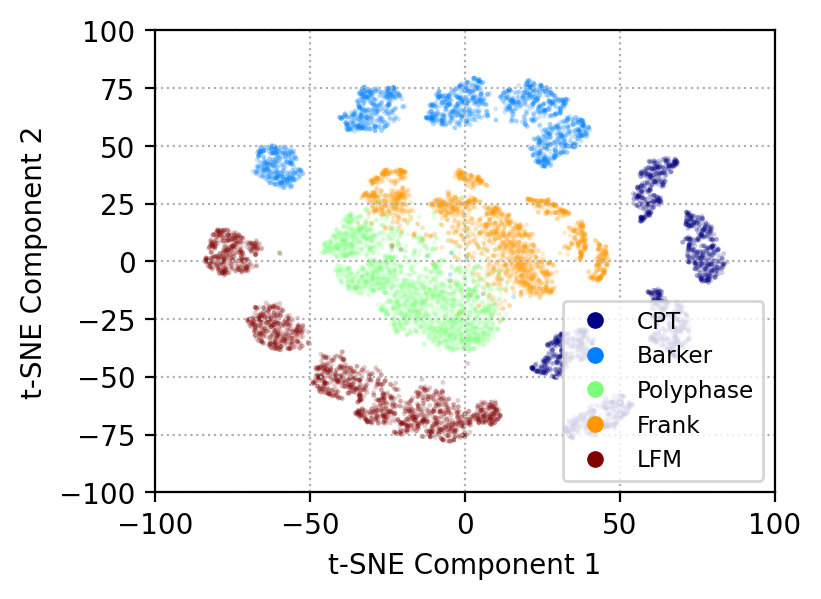}}
  \centerline{(a) Without SSL ($1$-shot)}\medskip
\end{minipage}
\hfill
\begin{minipage}[b]{0.495\linewidth}
  \centering
  \centerline{\includegraphics[width=1\linewidth]{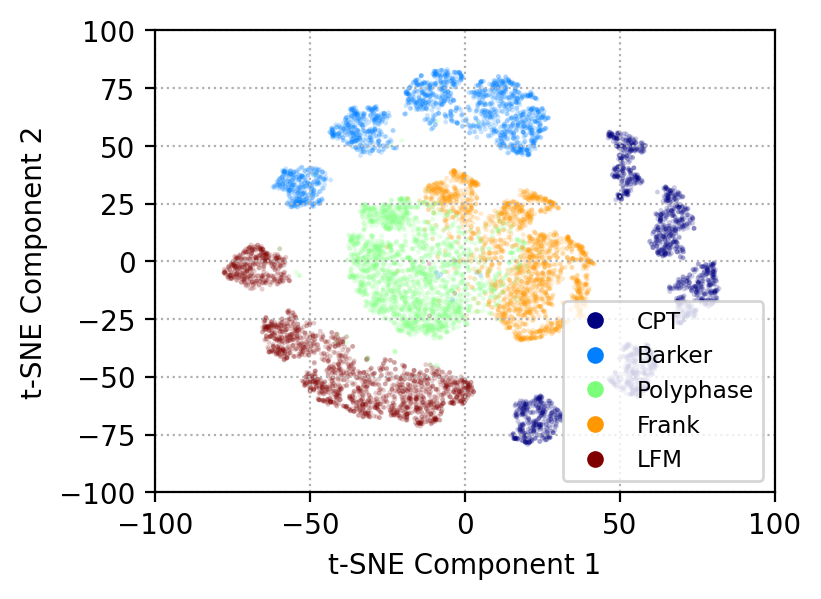}}
  \centerline{(b) Without SSL ($10$-shot)}\medskip
\end{minipage}
\hfill
\begin{minipage}[b]{0.495\linewidth}
  \centering
  \centerline{\includegraphics[width=1\linewidth]{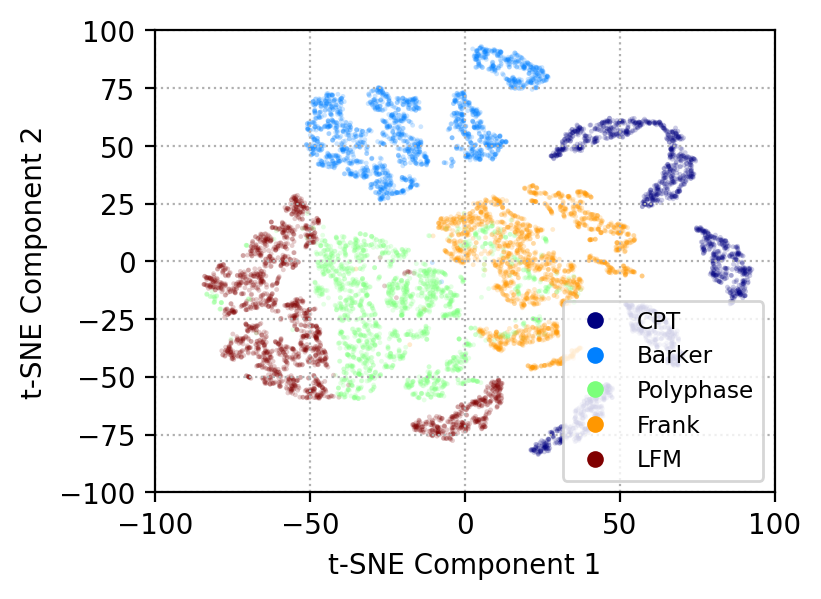}}
  \centerline{(c) RadChar-SSL (A, $0.7$)}\medskip
\end{minipage}
\hfill
\begin{minipage}[b]{0.495\linewidth}
  \centering
  \centerline{\includegraphics[width=1\linewidth]{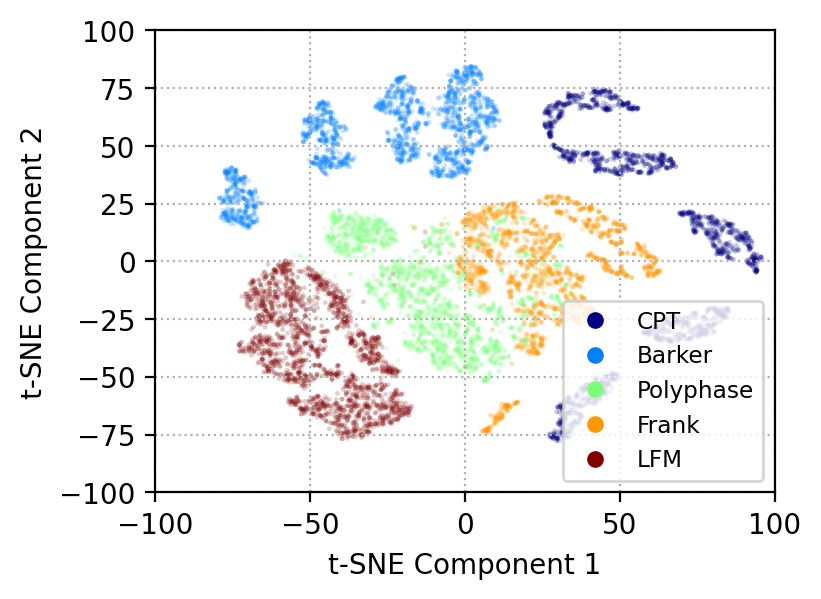}}
  \centerline{(d) DeepRadar (A, $0.4$)}\medskip
\end{minipage}
\hfill
\begin{minipage}[b]{0.495\linewidth}
  \centering
  \centerline{\includegraphics[width=1\linewidth]{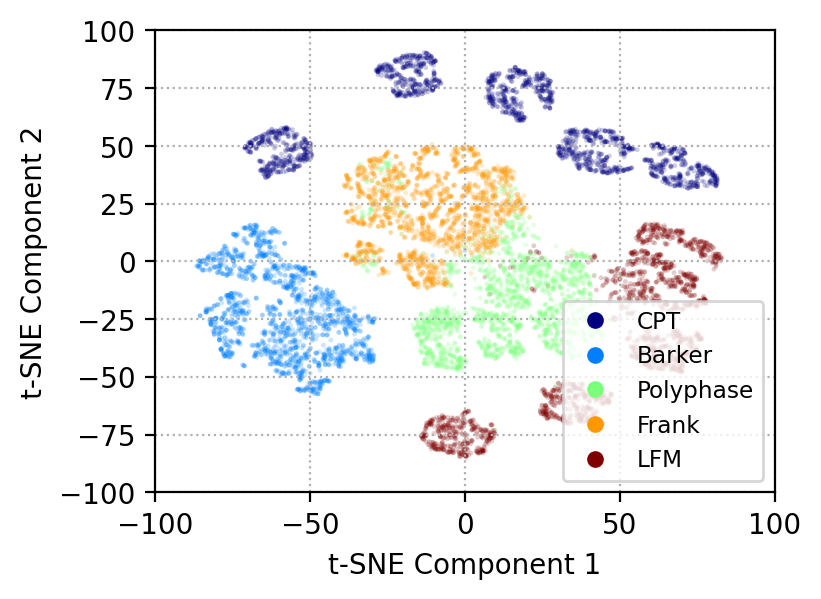}}
  \centerline{(e) RadarComm (D, $0.7$)}\medskip
\end{minipage}
\hfill
\begin{minipage}[b]{0.495\linewidth}
  \centering
  \centerline{\includegraphics[width=1\linewidth]{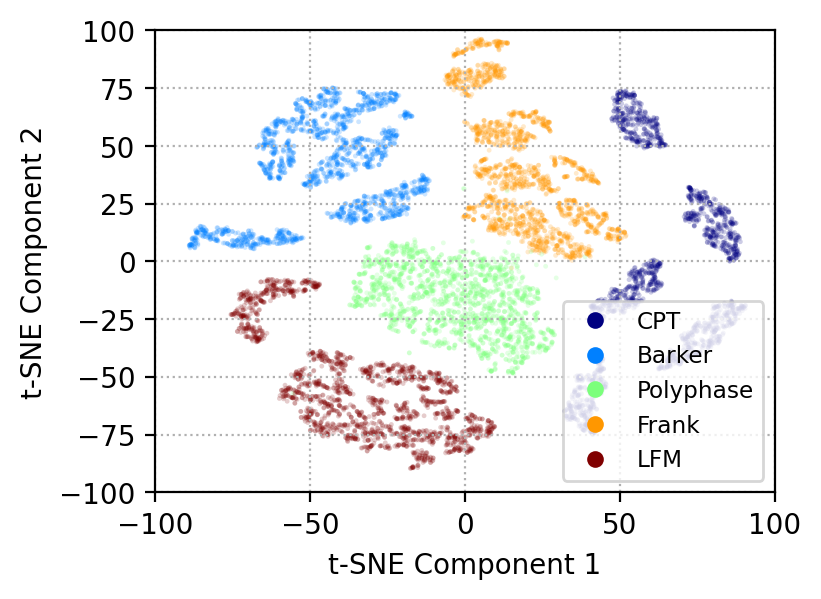}}
  \centerline{(f) RadioML (A, $0.1$)}\medskip
\end{minipage}
\hfill
\vspace{-22pt}
\caption{t-SNE results of $1$-shot fine-tuned ResNet1D encoder embeddings using the respective domain-optimal masking strategy ($S_{\text{m}}$, $R_{\text{m}}$). $10,000$ I/Q frames randomly selected from RadChar-Eval above $0$ \si{\deci\bel} SNR are shown.}
\label{fig:tsne}
\vspace{-10pt}
\end{figure}

To further evaluate the quality of SSL for RSR, we apply t-SNE to analyse how well models discriminate between radar classes during fine-tuning. For consistency, we first apply PCA to reduce the large pre-trained encoder embeddings to $50$ dimensions before using t-SNE to visualise the reduced embeddings in a 2D feature space. Our results (Fig. \ref{fig:tsne}) indicate that self-supervised pre-training with the optimal masking strategy enhances generalisation, allowing for effective adaptation from different RF domains. This is evident from the more compact clusters observed in Fig. \ref{fig:tsne}(c)–(f) compared to Fig. \ref{fig:tsne}(a) under the $1$-shot configuration. In contrast, ResNet1D trained without SSL exhibits a less distinct decision boundary between \textit{Polyphase} and \textit{Frank}, as both classes are polyphase-coded waveforms with similar intra-pulse characteristics \cite{huang2023multi}, making them difficult to distinguish. Furthermore, increasing the amount of annotated data for ResNet1D without SSL to a $10$-shot setting, as illustrated in Fig. \ref{fig:tsne}(b), only marginally improves the quality of inter-class separation. This observation is also reflected in Table \ref{table:massive_table}, where models fine-tuned using their respective domain-optimal masking strategy in a $1$-shot setting achieve performance comparable to models without SSL in a $10$-shot setting.

\vspace{-5pt}
\subsection{Ablation Study}
\label{ssec:ablation}
\vspace{-5pt}
We examine the impact of various design considerations for few-shot RSR. As discussed in Section \ref{ssec:discussion}, the choice of $S_{\text{m}}$ and $R_{\text{m}}$ influences SSL performance and, in turn, fine-tuning and test performance. Although no particular masking strategy consistently outperformed others across our experiments, we found that the masking ratio played a more important role in determining test performance (Table \ref{table:massive_table}). While a small batch size coupled with a low learning rate generally benefits fine-tuning, we observed that freezing the model weights for a few epochs provided a slight improvement (less than $5\%$) in test performance. This effect was consistent for models pre-trained on both in-domain and out-of-domain data. We also explored $\ell_2$ loss for pre-training, but we observed no meaningful improvements over $\ell_1$ loss in our experiments.
 

\vspace{-5pt}
\section{Conclusion}
\label{sec:conclusion}
\vspace{-5pt} 
In this paper, we introduced MSM as an effective SSL method for few-shot RSR. We also demonstrated the viability of RF domain adaptation for enhancing signal classification performance when no target domain data was used for pre-training. Our results show that by optimally designing the masking method during pre-training, fine-tuned models can achieve significant performance improvements, particularly in moderate to high SNR settings. This is demonstrated by ResNet1D, which achieved a boost in classification accuracy of $17.5\%$ when pre-trained on in-domain signals (RadChar-SSL), and $16.31\%$ when pre-trained on out-of-domain data (RadioML), compared to its $1$-shot baseline without SSL. In future work, additional downstream tasks may be explored to evaluate SSL and RF domain adaptation in a real-world setting.


\vfill
\pagebreak


\section{References} 
\vspace{-2\baselineskip}  

\scriptsize 
\bibliographystyle{IEEEbib}
\bibliography{refs}

\end{document}